\documentclass{article} 

\usepackage{algorithm}
\usepackage[noend]{algpseudocode}

\usepackage[final]{nips_2017}
\usepackage{tikz}
\usepackage{calc}
\usetikzlibrary{shapes.geometric}
\usepackage{amsmath}
\usepackage{bm}
\usepackage{graphicx}
\usepackage{caption}
\usepackage{subcaption}
\usepackage{mathtools}
\usepackage{wrapfig}
\usepackage[utf8]{inputenc} 
\usepackage[T1]{fontenc}    
\usepackage{hyperref}       
\usepackage{url}            
\usepackage{booktabs}       
\usepackage{amsfonts}       
\usepackage{nicefrac}       
\usepackage{microtype}      

\DeclarePairedDelimiterX{\infdivx}[2]{(}{)}{%
  #1\;\delimsize\|\;#2%
}

\newcommand{\infdivKL}{D_\text{KL}\infdivx}

\DeclareMathOperator{\Tr}{Tr}

\title{Information Maximizing Exploration with a Latent Dynamics Model}

\author{
  Trevor Barron, Heni Ben Amor\\
  Department of Computing, Informatics and Decision Systems Engineering\\
  Arizona State University\\
  United States\\
  \texttt{tpbarron,hbenamor@asu.edu}
  \And
  Oliver Obst\\
  Centre for Research in Mathematics, School of Computing, Engineering and Mathematics\\
Western Sydney University\\
Australia\\ 
\texttt{o.obst@westernsydney.edu.au}
}

\begin{document}

\maketitle

\begin{abstract}
All reinforcement learning algorithms must handle the trade-off between exploration and exploitation. Many state-of-the-art deep reinforcement learning methods use noise in the action selection, such as Gaussian noise in policy gradient methods or $\epsilon$-greedy in $Q$-learning. While these methods are appealing due to their simplicity, they do not explore the state space in a methodical manner. We present an approach that uses a model to derive reward bonuses as a means of intrinsic motivation to improve model-free reinforcement learning. A key insight of our approach is that this dynamics model can be learned in the latent feature space of a value function, representing the dynamics of the agent and the environment. This method is both theoretically grounded and computationally advantageous, permitting the efficient use of Bayesian information-theoretic methods in high-dimensional state spaces. We evaluate our method on several continuous control tasks, focusing on improving exploration.
\end{abstract}

\section{Introduction}

Model-free reinforcement learning (RL) has enjoyed significant success over the past few years. While model-based RL has not enjoyed the same successes, models have appealing properties that could potentially be leveraged to increase sample-efficiency, guide exploration, or enable high-level reasoning about internal actions. 

In this paper, we propose using a model to derive reward bonuses to accelerate a model-free RL technique. The key insight in our approach is that the dynamics model can be learned in the latent feature space of a value function. This builds upon existing work that formalizes an equivalence between a linear model and linear value estimate. We learn an actor-critic style policy and value function using standard model-free RL techniques, but simultaneously learn a 
Bayesian dynamics model (BNN) based on the \textit{latent feature} representation found by the final hidden layer of the value function approximator. The BNN takes as input the latent feature representation for a state and an action and predicts the latent feature representation for the state at the subsequent time step. Both the Bayesian model and the value function are linear in these features. The result is a compact representation of the task dynamics. In turn, the latent dynamics model can be used for intelligent, information theoretic exploration of the state space. The main contributions of this work include: 

	\begin{enumerate}
		\item Empirical results indicating that information maximizing exploration on latent features performs as good or better than an equivalent method on the true state space. 
        \item A Bayesian transition model learned in the latent feature space of a neural network value function capitalizing on already learned features that can be used to compute quantities such as information gain for exploration bonuses.
        \item An analysis of the relationship between the model distribution over latent features and the value distribution.
	\end{enumerate}
    
	The introduced method, which we call Information Maximizing Latent Exploration (IMLE), is both theoretically grounded and computationally advantageous. We evaluate our method on several continuous control, focusing on improving exploration.


\section{Methodology}
\label{sec:methodology}

	In this work we assume a finite-horizon discounted Markov decision process (MDP) defined by a tuple, $(\mathcal{S}, \mathcal{A}, \mathcal{P}, \mathcal{R}, \rho_0, \gamma)$, where $\mathcal{S} \subseteq \mathbb{R}^{n}$ is the state set, $\mathcal{A} \subseteq \mathbb{R}^m$ is the action set, $\mathcal{P} : \mathcal{S} \times \mathcal{A} \times \mathcal{S} \rightarrow \mathbb{R}_{\geq 0}$ is the state transition distribution, $\mathcal{R} : \mathcal{S} \times \mathcal{A} \rightarrow \mathbb{R}$ is the reward, $\rho_0$ is the start state distribution and $\gamma \in (0, 1]$ is the discount factor. We are interested in finding a stochastic policy $\pi_\theta : \mathcal{S} \times \mathcal{A} \rightarrow \mathbb{R}_{\geq 0}$ that maximizes the expected discounted reward, $\mathbb{E}_{\pi, \mathcal{P}} \left[\sum_{t=0}^T \gamma^t R(s_t, a_t)\right]$.
	
	\subsection{Relationship between model and value function}

	Several researchers have noted a relationship between transition models and value estimates \citep{parr2008analysis,sutton2008dyna}. Our approach extends previous work that studied the relationship between the solutions reached by a linear value function approximation and a one-step linear model. \citet{parr2008analysis} proved that the solution found by approximating a value function given an exact model and by solving for an exact value function given an approximate model are equivalent. This relationship motivates the decision to learn a dynamics model, not in the state space, but rather in the latent feature space of the value function. That is, we assume that features found to be useful for value estimation will also be useful for one-step model prediction. Furthermore, we argue that because the value estimate is linear in the latent features the model should be as well. Hence, we formulate the model as a linear Bayesian network.

	\subsection{A latent dynamics model}

		By logically decomposing the structure of a neural network into two components, first a feature learning component and second a linear approximation, we convert our problem into a form that closely mirrors the linear-case analysis equivalence by \citet{parr2008analysis}. A graphical representation of our network structure is shown in Figure \ref{fig:network}.

    \def\layersep{2.0cm}
	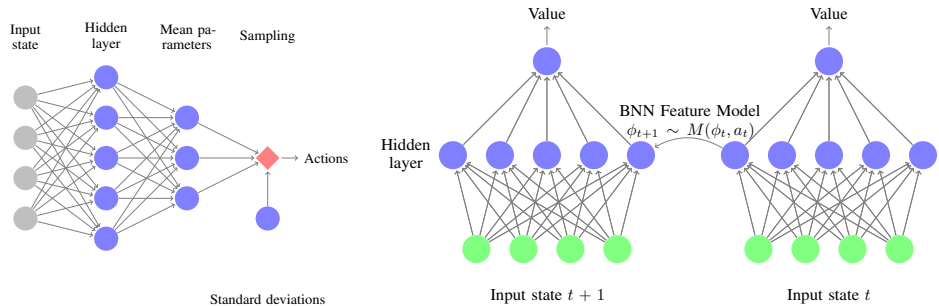
\begin{figure}
    \centering
	\resizebox{0.35\textwidth}{!}{
    \begin{tikzpicture}[shorten >=1pt,->,draw=black!50, node distance=\layersep]
      \tikzstyle{every pin edge}=[<-,shorten <=1pt]
      \tikzstyle{neuron}=[circle,fill=black!25,minimum size=17pt,inner sep=0pt]
      \tikzstyle{sampling}=[diamond,fill=black!25,minimum size=17pt,inner sep=0pt]
      \tikzstyle{input neuron}=[neuron, fill=gray!50];
      \tikzstyle{output neuron}=[neuron, fill=blue!50];
      \tikzstyle{hidden neuron}=[neuron, fill=blue!50];
      \tikzstyle{sample}=[sampling, fill=red!50];
      \tikzstyle{annot} = [text width=4em, text centered]
	  \tikzstyle{wideannot} = [text width=10em, text centered]

      \foreach \name / \y in {1,...,4}
          \node[input neuron] (I-\name) at (0,-\y) {};
      \foreach \name / \y in {1,...,5}
          \path[yshift=0.5cm]
              node[hidden neuron] (H-\name) at (\layersep,-\y cm) {};
      \foreach \name / \y in {1,...,3}
          \path[yshift=-0.5cm]
              node[hidden neuron] (M-\name) at (2*\layersep,-\y cm) {};
      \node[sample,pin={[pin edge={->}]right:Actions}, right of=M-2] (S) {};
      \node[hidden neuron] (D) at (3*\layersep,-4cm) {};
      \path (D) edge (S);
      \foreach \source in {1,...,4}
          \foreach \dest in {1,...,5}
              \path (I-\source) edge (H-\dest);
      \foreach \source in {1,...,5}
          \foreach \dest in {1,...,3}
             \path (H-\source) edge (M-\dest);
      \foreach \source in {1,...,3}
         \path (M-\source) edge (S);
      \node[annot,above of=H-1, node distance=1cm] (hl) {Hidden layer};
      \node[annot,left of=hl] {Input state};
      \node[annot,right of=hl] (ml) {Mean parameters};
      \node[annot,right of=ml] {Sampling};
      \node[wideannot,below of=D] {Standard deviations};
	\end{tikzpicture}
    }
	\resizebox{0.55\textwidth}{!}{
    \begin{tikzpicture}[shorten >=1pt,->,draw=black!50, node distance=\layersep]
	  \def\xnetoffseti{6}
      \def\xnetoffseth{5.5}
      \tikzstyle{every pin edge}=[<-,shorten <=1pt]
      \tikzstyle{neuron}=[circle,fill=black!25,minimum size=17pt,inner sep=0pt]
      \tikzstyle{sampling}=[diamond,fill=black!25,minimum size=17pt,inner sep=0pt]
      \tikzstyle{input neuron}=[neuron, fill=green!50];
      \tikzstyle{output neuron}=[neuron, fill=blue!50];
      \tikzstyle{hidden neuron}=[neuron, fill=blue!50];
      \tikzstyle{sample}=[sampling, fill=red!50];
      \tikzstyle{annot} = [text width=4em, text centered]
      \tikzstyle{wideannot} = [text width=10em, text centered]

      \foreach \name / \x in {1,...,4}
          \node[input neuron] (I1-\name) at (\x,0) {};

      \foreach \name / \x in {1,...,5}
          \path[xshift=-0.5cm]
               node[hidden neuron] (H1-\name) at (\x cm, \layersep) {};
               
      \node[output neuron,pin={[pin edge={->}]above:Value}, above of=H1-3] (O1) {};

      \foreach \source in {1,...,4}
		  \foreach \dest in {1,...,5}
              \path (I1-\source) edge (H1-\dest);

      \foreach \source in {1,...,5}
  	      \path (H1-\source) edge (O1);
          
      \foreach \source in {1,...,4}
		  \foreach \dest in {1,...,5}
              \path (I1-\source) edge (H1-\dest);

      \foreach \source in {1,...,5}
  	      \path (H1-\source) edge (O1);

      \foreach \name / \x in {1,...,4}
	      \path[xshift=\xnetoffseti cm]
    	      node[input neuron] (I2-\name) at (\x cm, 0) {};

      \foreach \name / \x in {1,...,5}
	      \path[xshift=\xnetoffseth cm]
              node[hidden neuron] (H2-\name) at (\x cm, \layersep) {};
               
      \node[output neuron,pin={[pin edge={->}]above:Value}, above of=H2-3] (O2) {};

      \foreach \source in {1,...,4}
		  \foreach \dest in {1,...,5}
              \path (I2-\source) edge (H2-\dest);

      \foreach \source in {1,...,5}
  	      \path (H2-\source) edge (O2);
          
      \foreach \source in {1,...,4}
		  \foreach \dest in {1,...,5}
              \path (I2-\source) edge (H2-\dest);

      \foreach \source in {1,...,5}
  	      \path (H2-\source) edge (O2);

      \path [bend right] (H2-1) edge (H1-5);
      \node[wideannot,above left of=H2-1, xshift=-0.25cm, node distance=1cm] (hl1) {BNN Feature Model\\ $\phi_{t+1} \sim M(\phi_t, a_t)$};

      \node[annot,left of=H1-1, node distance=1cm] (hl1) {Hidden layer};
      \node[wideannot,below of=I1-3,xshift=-0.5cm,yshift=1cm] {Input state $t+1$};
      \node[wideannot,below of=I2-3,xshift=-0.5cm,yshift=1cm] {Input state $t$};

	\end{tikzpicture}
    }
    \caption{The network structures used to approximate the policies in the vector and high-dimensional cases (left). A Bayesian network is trained in the latent feature space of a value function approximator (right).}
    \label{fig:network}
	\end{figure}

\begin{figure}[!htb]
        \centering
    	\begin{subfigure}[t]{0.5\textwidth}
        	\centering
            \includegraphics[width=\linewidth]{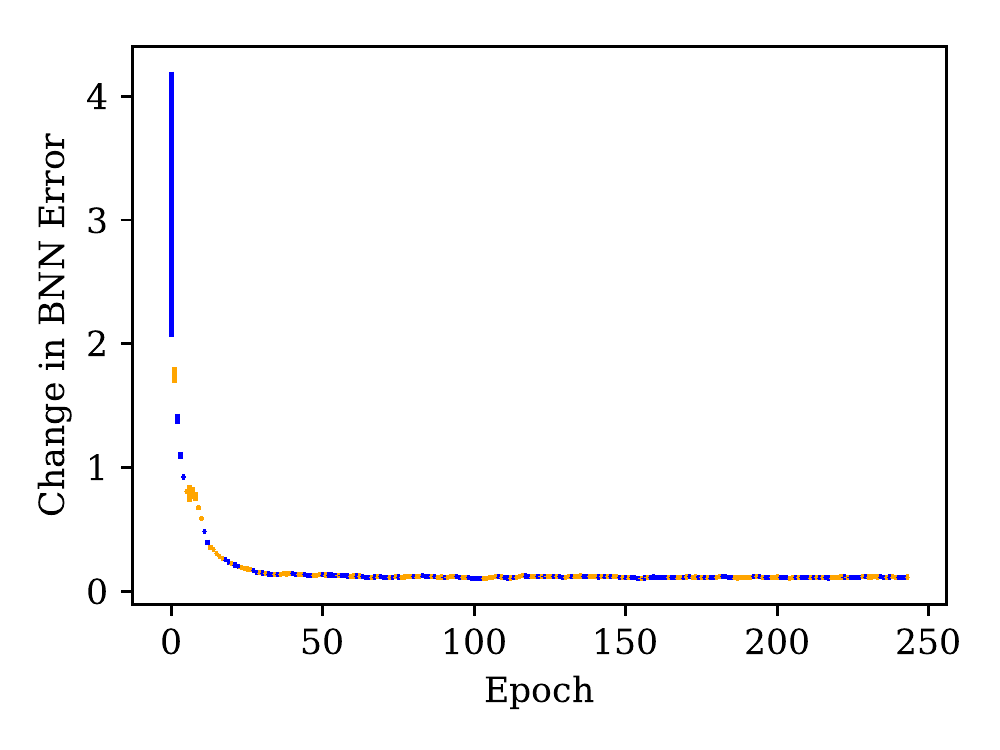}
	  		\caption{Dense Walker2D}
    	\end{subfigure}%
    	~ 
    	\begin{subfigure}[t]{0.5\textwidth}
        	\centering
	  		\includegraphics[width=\linewidth]{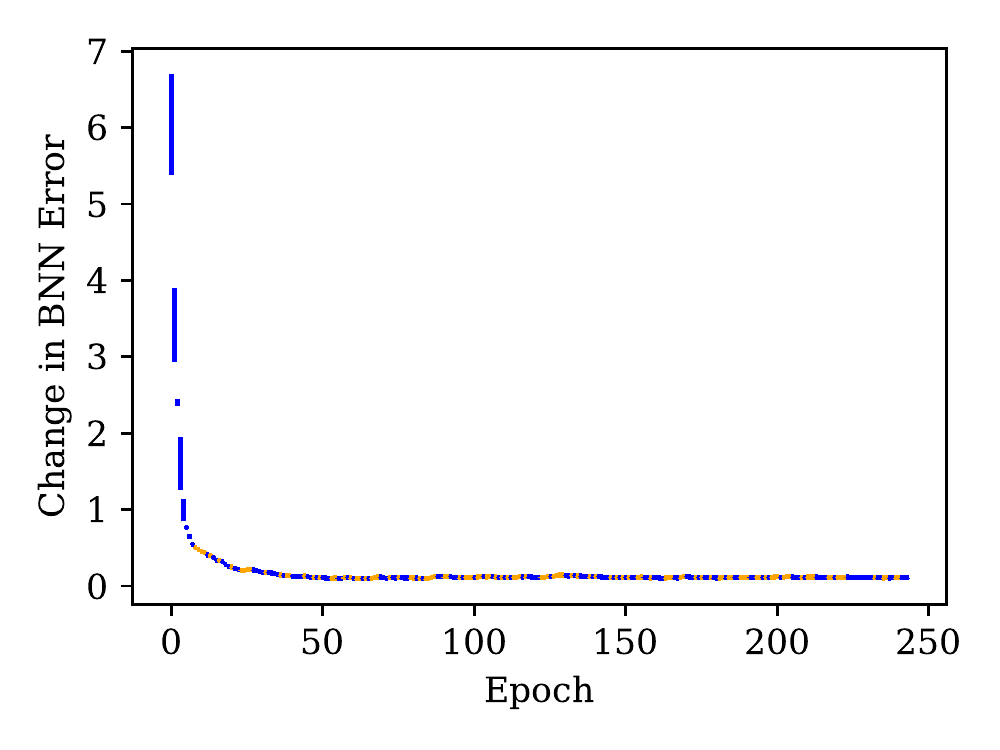}
	  		\caption{Dense Ant}
    	\end{subfigure}%

        \begin{subfigure}[t]{0.5\textwidth}
        	\centering
	  		\includegraphics[width=\linewidth]{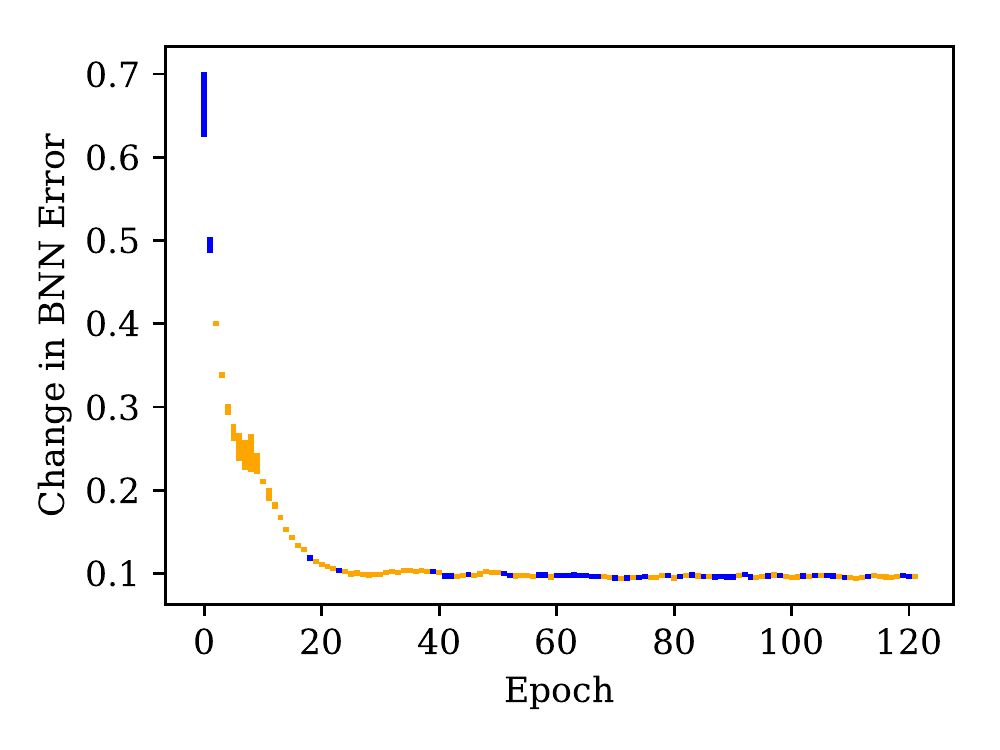}
	  		\caption{Sparse Mountain Car}
		\end{subfigure}%
        ~
		\begin{subfigure}[t]{0.5\textwidth}
        	\centering
	 		\includegraphics[width=\linewidth]{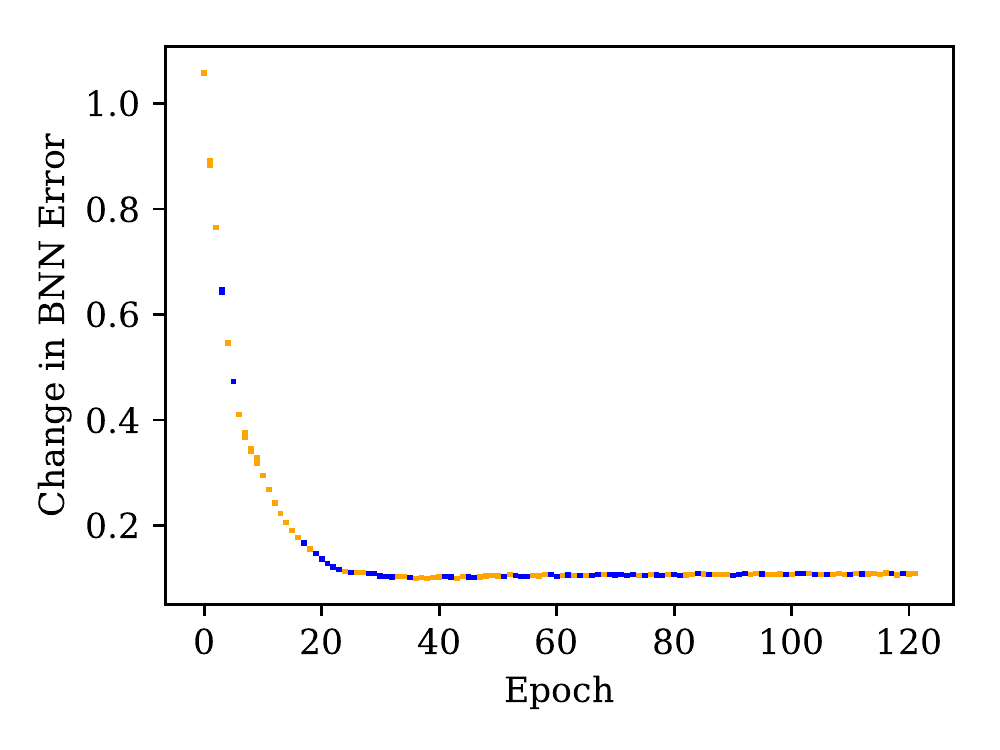}
	 		\caption{Sparse Acrobot}
        \end{subfigure}
        \caption{Blue bars represent BNN improvement during a value function update, orange bars represent BNN accuracy decline. Early in training updating the value function reduces BNN model error. This empirical finding supports the theoretical model-value relationship and motivates learning the model in the feature space of the value function.}
        \label{fig:bnn_errors}
		\end{figure}

		We alternate between approximating the value function and a linear Bayesian dynamics model given the current value function's latent features. Using value function features as basis for model learning results in non-stationary features. In practice we do not find this to be a problem and observe that the latent dynamics model converges rapidly. In fact, we empirically find that updating the value function often decreases the error of the dynamics model, especially early in training (see Figure \ref{fig:bnn_errors}). This finding highlights the relationship between features for value estimation and features for prediction and validates our approach.
        
		Employing a Bayesian dynamics model provides additional useful information including an uncertainty measure regarding a prediction and a means of computing information gained by visiting a state (through KL divergence). A single forward step can be expressed as $\Pr(\phi_{V_\theta}(\hat{s}_{t+1}) \mid \phi_{V_\theta}(s_t), a_t;\omega)$ where $\phi_{V_\theta}(s_t)$ is the output from the final hidden layer of the value function (pre-activations), $V_\theta$, and $\omega$ are the parameters of the model sampled from a random variable $\bm{\Omega}$. A distribution over possible models is maintained through a prior $\Pr(\omega)$, which we choose to be Gaussian.

		Operating in the latent space also provides the means to obtain a measure of uncertainty not only over our state transition but also over the future value approximation. Note that since the value estimate is simply a linear transformation of the latent features, the output distribution of the model can be transformed to represent a distribution over the value estimate. The value function approximation remains a deterministic linear transform but a distribution over the \textit{inputs} is given by the dynamics model. If the probability of a latent encoding of the value function given a state, $s$, is drawn from a normal distribution $\Pr(\bm{L}) = \Pr(\phi_{V_\theta}(s_t)) \sim \mathcal{N}(\mu_s,\,\sigma^{2}_s)$, as is the case with our linear Bayesian network, then a distribution over the value function itself is given by:
        
		\begin{align}
			\mathrm{E}(\bm{V}) &= W\mathrm{E}(\bm{L}) + \bm{b} \\
			\mathrm{Var}(\bm{V}) &= W\mathrm{Cov}(\bm{L})W^T\text{,}
		\end{align}
		
		where $W$ and $\bm{b}$ are the weights and biases of the final linear layer. Accordingly, the distribution of the value estimate is $\Pr(V_\theta(s_t)) \sim \mathcal{N}(\mathrm{E}(\bm{V}), \mathrm{Var}(\bm{V}))$. Hence by learning a Bayesian dynamics model in latent feature space we get an uncertainty measure of the value function for free without incorporating any weight uncertainty into the value function itself.
	
        
	\subsection{Incentivizing exploration with reward bonuses and intrinsic motivation}

		In this work we focus on exploration and evaluate a method akin to Variational Information Maximizing Exploration (VIME) \citep{houthooft2016vime} in the latent space. Reinforcement learning requires a careful balance between exploration and exploitation. One approach is to provide reward bonuses in order to incentivize an agent to take actions even in the absence of well-shaped rewards from the environment. 
    
	This exploration strategy is based on the idea that an agent should behave in a manner to reduce its uncertainty regarding the environment dynamics. Bayesian models handle the trade off between exploration and exploitation naturally. In VIME, optimizing a Bayesian model is made tractable with variational inference and reward bonuses are derived from the amount of information gained by visiting a state. In particular, an intrinsic reward is derived for reductions in model entropy. This encourages the agent to take actions that provide maximal information about the environment. Following the notation in \citet{houthooft2016vime}, this quantity can be expressed formally as the information gain of observing a future state $s_{t+1}$, taking action $a_t$, with history $\zeta_t$,
    \begin{align}
    	I(S_{t+1}; \Omega \mid \zeta_t, a_t) = \mathbb{E}_{s_{t+1} \sim \mathcal{P}(\cdot | \zeta_t, a_t)} [\infdivKL{p(\omega | \zeta_t, a_t, s_{t+1})}{p(\omega | \zeta_t)}]\text{.}
    \end{align}
    
    The reward bonus is then formulated as,
    \begin{align}
    	r'(s_t, a_t, s_{t+1}) = r(s_t, a_t) + \eta \infdivKL{p(\omega | \zeta_t, a_t, s_{t+1})}{p(\omega | \zeta_t)}\text{,}
    \end{align}
    
    where $\eta$ is a parameter that controls the trade-off between exploration and exploitation.
    
    In practice, computing the distributions within the KL divergence are usually intractable. Accordingly, we approximate this distribution over the parameters given the agent's experience using variational inference \citep{hinton1993keeping}. This approach minimizes $\infdivKL{q(\omega; \theta)}{p(\omega)}$ where $q(\omega; \theta)$ is a probabilistic model parameterized by $\theta$. The information gain for visiting a state is then approximated using the KL divergence between the parameters of the model before and after observing $s_{t+1}$,  $\infdivKL{q(\omega; \theta_{t+1})}{q(\omega; \theta_{t})}$ and the reward bonus becomes,
    \begin{align}
    	r'(s_t, a_t, s_{t+1}) = \underbrace{r(s_t, a_t)}_{\text{Environ. reward}} + \eta \underbrace{\infdivKL{q(\omega; \theta_{t+1})}{q(\omega; \theta_{t})}}_{\text{Intrinsic reward}}\text{.}
    \end{align}
    
    As in VIME we model the Bayesian network as a fully-factored Gaussian distribution, which makes it possible to approximate the KL divergence with a single second order gradient step.
   
    
    The algorithm pseudo-code for IMLE is specified in Algorithm \ref{alg:latent_rl}. For more details on VIME specifics we refer readers to the original paper \citep{houthooft2016vime}.

	\begin{algorithm}
		\caption{Information maximizing exploration with a latent Bayesian dynamics model}
  		\label{alg:latent_rl}
		\begin{algorithmic}[1]
		\State $\text{Initialize policy}\, \pi_0, \text{Bayesian model}\, \mathit{M}, \text{replay memory}\, \mathit{R}$
		\For{$\text{epoch} = 0, 1, 2, \ldots \text{until convergence}$}
			\For{\text{time step in epoch}}
				\State \text{Behave according to policy} $a_t = \pi_{\theta_p}(s_t)$
				\State \text{Save transition} $(s_t, a_t, s_{t+1})$\,\text{to replay memory,}\,$R$
                \State \text{Project} $l_t = \phi(s_t)$ and $l_{t+1} = \phi(s_{t+1})$
                \State \text{Compute} $\infdivKL{q(\omega; \theta_{t+1})}{q(\omega; \theta_{t})}$ by approximation $\nabla_{\theta}^T H^{-1} \nabla_{\theta}$
				\State \text{Compute rewards} $r'(s_t, a_t, s_{t+1}) = r(s_t, a_t) + \infdivKL{q(\omega; \theta_{t+1})}{q(\omega; \theta_{t})}$
			\EndFor
			\State \text{Draw samples, $S$, from replay,} $R$
			\State \text{Project samples to latent representation} $l_s = \phi(s)$
			\State \text{Minimize variational loss of Bayesian dynamics model, $M$, on one-step latent feature}
            \State \text{\quad dynamics, $l_{s_{t+1}} \sim M(l_{s_t}, a_t)$}
			\State \text{Update policy,} $\pi$ \text{, given augmented rewards with standard RL method}
		\EndFor
  		\end{algorithmic}
	\end{algorithm}

	\subsection{A relationship between model-based information gain-based intrinsic motivation and Bayesian Q-Learning in the Linear Case}

	We now discuss a relationship between intrinsic rewards based on information gain of a dynamics model and information gain of a value estimate when both the model and value estimate are linear in the feature representation. Specifically, we find that if the Bayesian dynamics model represents the true distribution over the future latent features, then the intrinsic reward bonus derived from model-based information gain is greater than or equal to a bonus derived from information gain regarding the $Q$-value estimate. We assume $M(s_t, a_t) \stackrel d= \phi^*_{t+1}$, where $M(s_t, a_t)$ is the distribution over latent features predicted by the model at time $t+1$ and $\phi^*_{t+1}$ represents the true distribution over the latent features at time $t+1$.
    
    The KL divergence of the dynamics model is computed as a sum of the KL divergence over each parameters' old and new distribution after observing a new sample. Specifically, let $w_1 \sim \mathcal{N}(\mu_{w_1}, \sigma_{w_1}^2)$ and $b_1 \sim \mathcal{N}(\mu_{b_1}, \sigma_{b_1}^2)$. We then let the agent take an action $a$ and transition to state $s$. Updating the dynamics model with this new information gives $w_2 \sim \mathcal{N}(\mu_{w_2}, \sigma_{w_2}^2)$ and $b_2 \sim \mathcal{N}(\mu_{b_2}, \sigma_{b_2}^2)$. Then, the information gain according to the model is $IG_{model} = \infdivKL{w_1}{w_2} + \infdivKL{b_1}{b_2}$. Given the corresponding parameter distributions, we find the output distributions for sample, $s$, before and after observing $s$ to be $o_1 \sim \mathcal{N}(x\mu_{w_1} + \mu_{b_1}, x^2\sigma_{w_1}^2 + \sigma_{b_1}^2)$ and $o_2 \sim \mathcal{N}(x\mu_{w_2} + \mu_{b_2}, x^2\sigma_{w_2}^2 + \sigma_{b_2}^2)$. This follows because the prior is Gaussian and the model is linear.
    
    Moreover, since the KL divergence is preserved under linear transformation (proof in Appendix), the KL divergence of the output distribution is identical to the KL divergence of the $Q$-value distribution before and after observing $s$. Specifically, assume a mean, $\mu_{m_1}$, and standard deviation, $\sigma_{m_1}$, defined by the output of the dynamics model before observing $s$ and a new distribution defined by $\mu_{m_2}$ and $\sigma_{m_2}$ after observing $s$. We then examine the KL divergence between $m_1(x) = \mathcal{N}(\mu_{m_1}, \sigma_{m_1}^2)$ and $m_2(x) = \mathcal{N}(\mu_{m_2}, \sigma_{m_2}^2)$. Since we assume a network structure such that the value estimate is only a linear transformation of the output of the dynamics model, it too defines a Gaussian distribution. Given parameters $w$ and $b$ of the linear transformation, these transformed distributions, $q_1(x) = \mathcal{N}(w\mu_{m_1} + b, w^2\sigma_{m_1}^2)$ and $q_2(x) = \mathcal{N}(w\mu_{m_2} + b, w^2\sigma_{m_2}^2)$ represent the distribution of $Q(s, a)$ before and after observing $s$. The KL divergence of the output of dynamics model is equivalent to the KL divergence of the $Q$-value estimate. 
    
    Empirically, by sampling, we find $\infdivKL{q_1}{q_2} = \infdivKL{m_1}{m_2} \leq IG_{model}$. (It may be possible to prove generally that $\infdivKL{m_1}{m_2} \leq IG_{model}$ when both are linear in the feature representation though thus far the authors have been unable to show this in a general case.) Intuitively, this means that incentivizing exploration to areas of high value uncertainty, possibly derived from a Bayesian Q-network, provides a weaker intrinsic reward than incentivizing exploration to areas of high model uncertainty. Further analysis of when the divergence of the model and the divergence of the $Q$-value estimate differ is left to future work.
    

\section{Experimental Results}
\label{sec:result}
    
	We focus our study on actor-critic policy gradient methods where both policy and value function approximators are learned. While some methods have merged the policy and value function into a single function approximator with two heads we instead learn separate models for each in order to align with the theory regarding model-value equivalence as closely as possible \citep{mnih2016asynchronous,schulman2017ppo}. In this work we use a Proximal Policy Optimization (PPO) method as a baseline \citep{schulman2017ppo}. We evaluate our method on continuous control benchmark tasks from the OpenAI Gym based on Box2D and the Pybullet simulator \citep{openaigym,coumans2017pybullet,Box2D}. We compare our method to PPO with and without VIME. All experiments are averaged over three random seeds. In general we find that IMLE performs slightly better than VIME on continuous control tasks. 
    
    \begin{figure}[!htb]
        \centering
 		\includegraphics[width=0.25\linewidth]{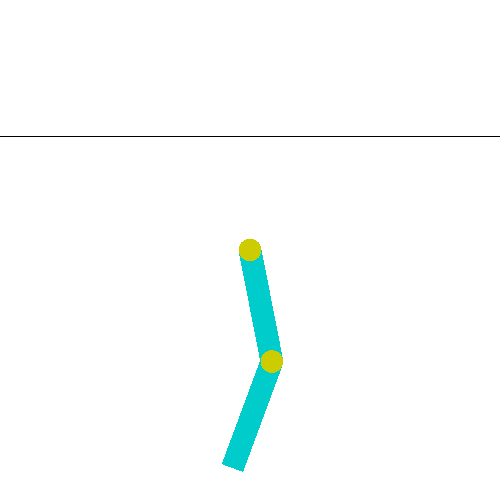}
 		\label{fig:acrobot_example}
  		\includegraphics[width=0.25\linewidth]{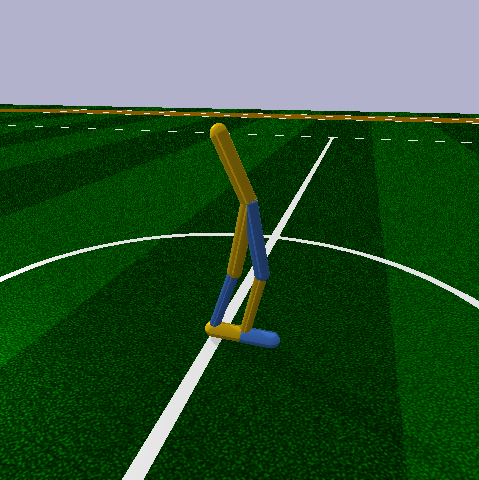}
  		\label{fig:walker2d_example}
        \caption{Example renders of the Acrobot and Walker2D tasks.}
	\end{figure}
        
	\subsection{Implementation Details}
    
    	The IMLE and VIME, BNN updates and intrinsic reward generation is integrated into the PPO algorithm. The agent interacts with the environment for a defined number of steps, $K$. Each transition is added to a FIFO replay memory. After $K$ steps have passed the RL policy and value function are updated according to the PPO update algorithm. If the replay memory is of sufficient size, the BNN is then updated as well. Once the BNN has been updated once, it is used to derive an intrinsic reward on for each transition experienced by the agent. The intrinsic reward given to the agent is normalized by the median of the 10 previous KL divergence epoch means and then scaled by $\eta$. In IMLE, the encoded representation is taken before applying any non-linearity but is then normalized using a running mean and standard deviation. A full table of hyperparameters is in the Appendix. Both the policy and value function are represented as neural networks. In all tasks the policy had two hidden layers of 64 nodes and the value function had a two hidden layers of size 32. The latent feature size is therefore 32 as well.
    
    \subsection{Tasks with Sparse Rewards}
    
\begin{figure}[!htb]
        \centering
    	\begin{subfigure}[t]{0.5\textwidth}
        	\centering
            \includegraphics[width=\linewidth]{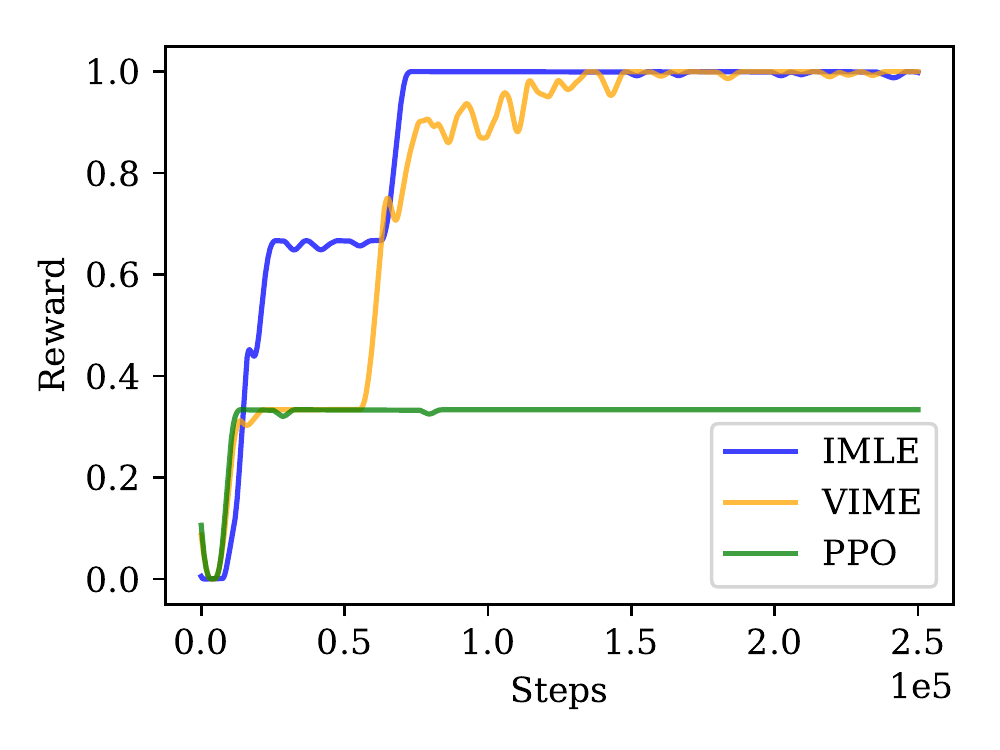}
            \caption{Sparse Acrobot}\label{fig:acrobot_low_d}
    	\end{subfigure}%
    	~ 
    	\begin{subfigure}[t]{0.5\textwidth}
        	\centering
            \includegraphics[width=\linewidth]{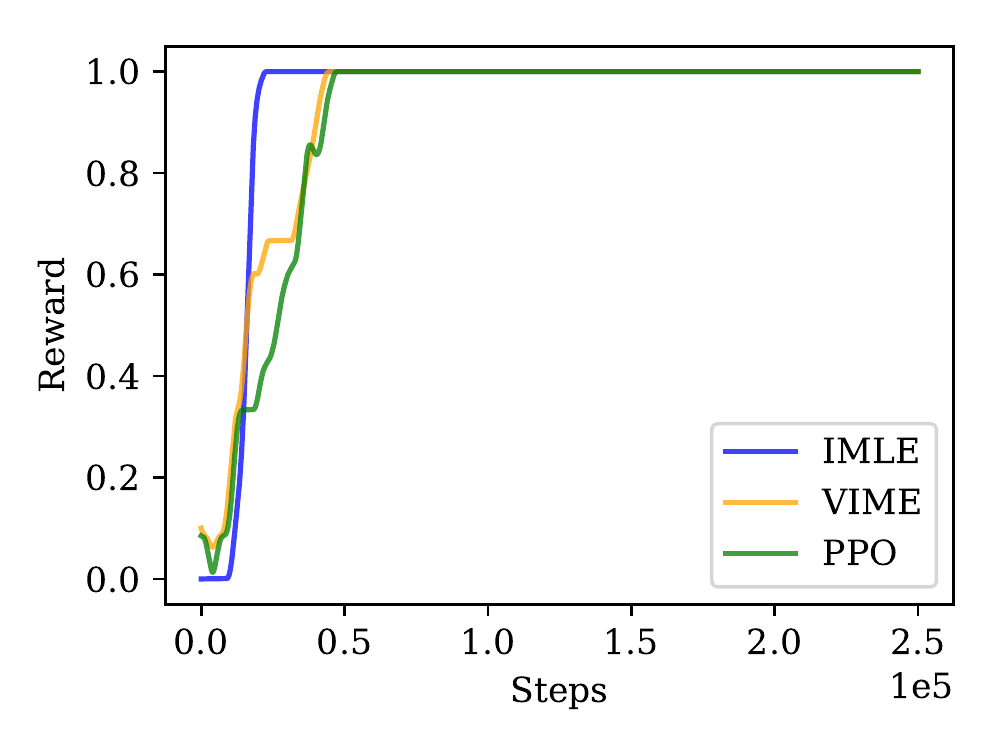}
            \caption{Sparse Mountain Car}\label{fig:mountain_car_low_d}
    	\end{subfigure}%
        \caption{We evaluate IMLE vs VIME and PPO on the Acrobot and MountainCar tasks with sparse rewards.}
		\end{figure}

 		To examine how effectively our method incentivizes exploration we modify the Acrobot ($\mathcal{S}\in \mathbb{R}^{6}, \mathcal{A}\in \mathbb{R}$) and MountainCar ($\mathcal{S}\in \mathbb{R}^{2}, \mathcal{A}\in \mathbb{R}$) environments to have very sparse rewards, therefore requiring an efficient exploration strategy \citep{sutton1996generalization}. A reward of positive one is given when the goal state is reached and the reward is zero everywhere else. The Acrobot environment is also modified to have continuous torque input with a maximum magnitude of one. Figures \ref{fig:acrobot_low_d} and \ref{fig:mountain_car_low_d} show the performance of IMLE versus VIME and PPO. Without intrinsic rewards, PPO is unable to solve the Acrobot task consistently. While both IMLE and VIME are able to solve the task, IMLE reaches the optimal reward of one more rapidly. Note that the unusual step-like nature of the IMLE learning curve is not reflective of the algorithm, instead it is a remnant of averaging multiple training runs. In the MountainCar environment all three algorithms solve the task but IMLE solves the task most rapidly.
        
	\section{Tasks with Well-Shaped Rewards}
	
    	In addition to the Acrobot and MountainCar tasks, we also test our method on continuous control tasks in the Pybullet simulator. We run experiments on Walker2D ($\mathcal{S}\in \mathbb{R}^{22}, \mathcal{A}\in \mathbb{R}^6$) and Ant ($\mathcal{S}\in \mathbb{R}^{28}, \mathcal{A}\in \mathbb{R}^8$). 
        
\begin{figure}[!htb]
        \centering
    	\begin{subfigure}[t]{0.5\textwidth}
        	\centering
            \includegraphics[width=\linewidth]{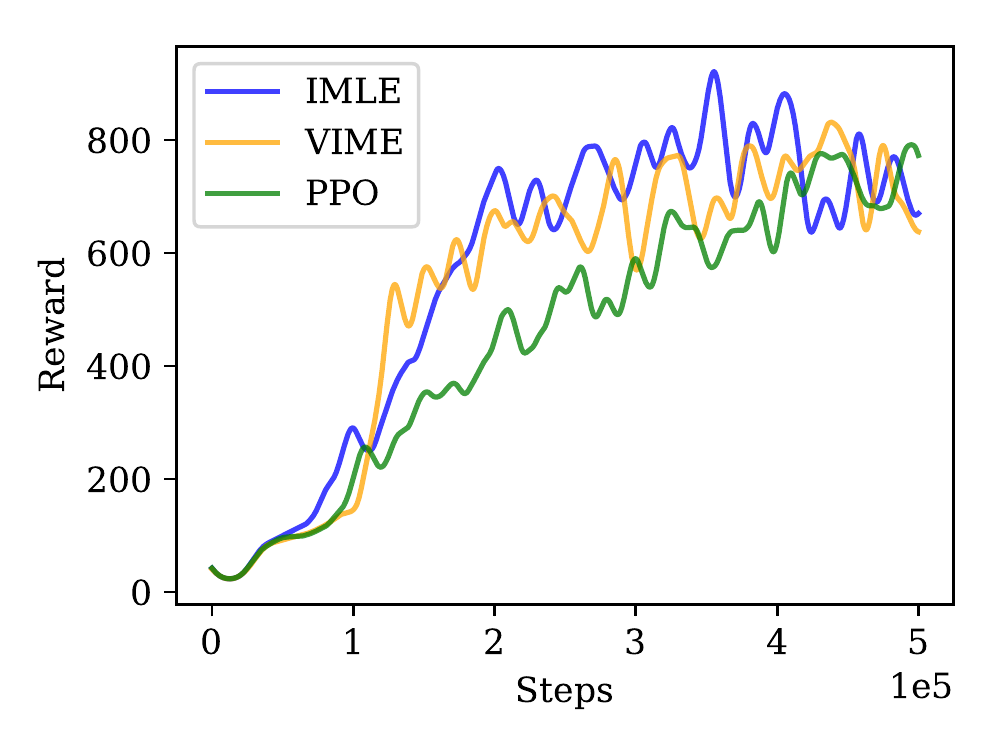}
            \caption{Dense Walker2D}\label{fig:walker2d_low_d}
    	\end{subfigure}%
    	~ 
    	\begin{subfigure}[t]{0.5\textwidth}
        	\centering
            \includegraphics[width=\linewidth]{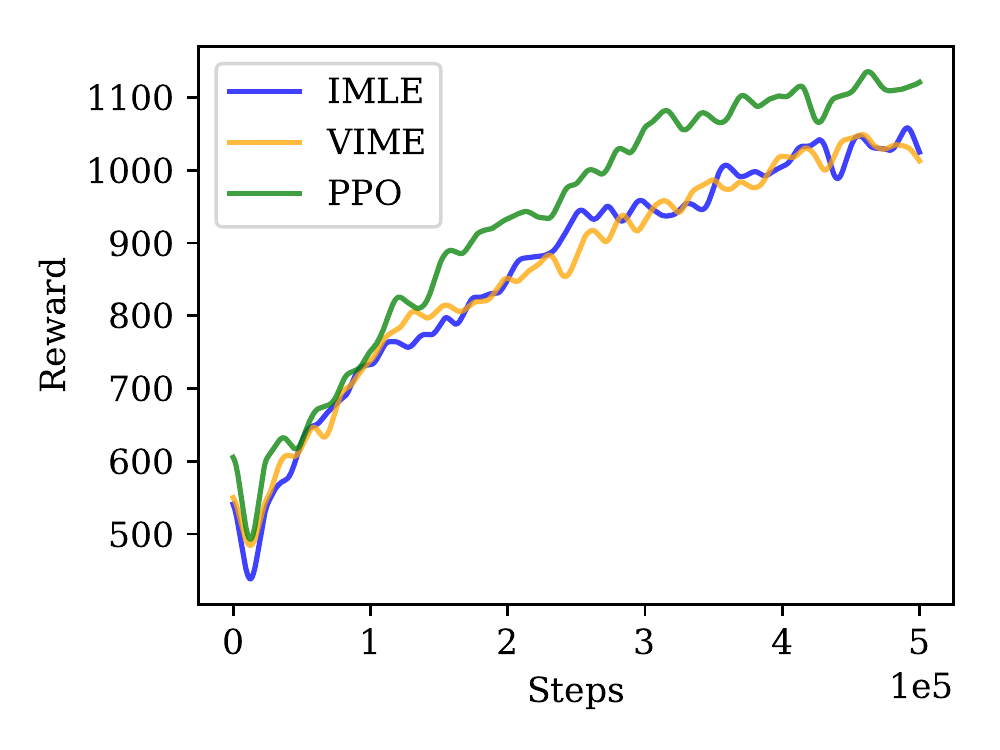}
            \caption{Dense Ant}\label{fig:ant_low_d}
    	\end{subfigure}%
        \caption{We evaluate IMLE vs VIME and PPO on the Walker2D and Ant tasks with dense rewards.}
		\end{figure}

        In these tasks we do not modify the reward. Instead we wish to observe how IMLE fares in tasks with carefully shaped rewards. Interestingly, we find that even with well-shaped rewards IMLE can still provide improvement and, moreover, that it can outperform VIME (Figure \ref{fig:walker2d_low_d}). In the Walker2D task both IMLE and VIME out-pace vanilla PPO with IMLE slightly above VIME. This is not always the case, though, as additional exploration can also reduce performance. In the Ant environment, both IMLE and VIME marginally under perform relative to vanilla PPO (Figure \ref{fig:ant_low_d}).

   \subsection{Computational Considerations}
    	
        Bayesian models are known to have high computational requirements as a result of the need to make multiple predictions in order to estimate evidence lower bound. Operating in the latent feature space also provides appealing computational improvements over VIME as the observation space grows. When the dynamics model is trained directly on state inputs, more parameters are needed in the BNN for good accuracy. 
        
        For example, consider the Walker2D task, which has $\mathcal{S}\in \mathbb{R}^{22}$ and $\mathcal{A}\in \mathbb{R}^6$. Then the BNN used in VIME has an input size of 28 (concatenated state and action vectors), two layers of size 32, and an output size of 22. If we draw 10 samples to estimate the variational lower bound, then the number of multiplications required for this operation is $(n_\text{input} \cdot n_\text{out1} + n_\text{out1} \cdot n_\text{out2} \cdot n_\text{out2} \cdot n_\text{pred}) \cdot n_{samples} = (28 \cdot 32 + 32 \cdot 32 + 32 \cdot 22) \cdot 10 = 26240$.
        
        A latent model fares better as the dimensionality of the input increases as the encoding occurs only once and the sampling is then done on a linear model. We assuming a value function with two hidden layers each of size 32 as is used in our work. Then the number of multiplications is $n_\text{in} \cdot n_\text{out1} + n_\text{out1} \cdot n_\text{out2} \cdot (n_\text{modelin} \cdot n_\text{modelout}) \cdot n_\text{samples} = 28 \cdot 32 + 32 \cdot 32 + (38 \cdot 32) \cdot 10 = 14080$.
        
        That is, compared to VIME, the IMLE update requires approximately half the number of multiplications to estimate the output distribution and BNN loss. This can be interpreted as a hybrid approach that combines the benefit of neural network feature extraction and Bayesian methods. In general, as the complexity of the state representation grows the deterministic encoding results in an increasingly more efficient BNN update.

\section{Related Work}
\label{sec:related}

	This work received inspiration most directly from VIME \citep{houthooft2016vime} and the analysis of linear models and linear value function approximation by \citet{parr2008analysis}. VIME provides intrinsic rewards based on information gain computed from a Bayesian dynamics model. \citet{parr2008analysis} developed the theory equating the solutions of linear fixed point value function methods and linear models. 
    
	A large corpus of work has been devoted to developing exploration strategies for reinforcement learning agents. The idea of working in a lower dimensional space is not new. It is common in robotics applications to perform dimensionality reduction to increase convergence rate. \citet{lange2012autonomous} use an autoencoder to extract features on which a policy is trained. \citet{luck2016sparse} employ a linear probabilistic dimensionality reduction that is learned during training. Both of these methods require learning additional features for the dimensionality reduction, as opposed to our method which operates on existing features being learned by the value function. 
    
    Several published methods use prediction for exploration. \citet{DBLP:journals/corr/StadieLA15} use an autoencoder to derive novelty bonuses based on prediction error of a latent state encoding. While this work does prediction in the latent encoding, it does not take advantage of features learned by the Q-network and learns policies only in discrete action spaces. \citet{pathak2017curiosity} recently published a method that uses intrinsic rewards based on a forward model to provide reward bonuses based on high model error and successfully learn policies in environments with sparse or no rewards.
    
    None of the aforementioned works have considered sharing features between a value function and dynamics model or use a latent feature-level dynamics model for exploration.

    

\section{Conclusion}
\label{sec:conclusion}

	This work presented a method for learning a probabilistic model, which can be used to derive intrinsic reward bonuses, in the learned feature space of a value function. We show results indicating that our method, IMLE, achieves performance comparable or better than VIME.
    
    Since operating in the latent space unties the dimensionality of the input observation from the dimensionality of the Bayesian model, we believe that IMLE has the potential to scale to much higher dimensional problems including pixel observations. 
    
    Additionally, the improvement in model error on value function updates indicate a very close practical model-value relationship. Accordingly, we also believe it is worthwhile to investigate updating a value function using the model \textit{gradient} (as opposed to model-based RL updates). 
    
    Finally, while the proposed network structure has theoretical foundations, it is logical to wonder whether a similar approach is possible in the feature space of the policy or a joint policy-value network as in A3C and some PPO implementations \citep{mnih2016asynchronous,schulman2017ppo}. A joint structure is appealing as it would provide a means to perform model-based RL updates as well. Our first attempts with such a structure did not perform as well as when learning the model only in the feature space of the value function though we do believe there is future work to be done in this area.


\bibliographystyle{plainnat}
\bibliography{nips2017_conference}

\begin{thebibliography}{14}
\providecommand{\natexlab}[1]{#1}
\providecommand{\url}[1]{\texttt{#1}}
\expandafter\ifx\csname urlstyle\endcsname\relax
  \providecommand{\doi}[1]{doi: #1}\else
  \providecommand{\doi}{doi: \begingroup \urlstyle{rm}\Url}\fi

\bibitem[Brockman et~al.(2016)Brockman, Cheung, Pettersson, Schneider,
  Schulman, Tang, and Zaremba]{openaigym}
Greg Brockman, Vicki Cheung, Ludwig Pettersson, Jonas Schneider, John Schulman,
  Jie Tang, and Wojciech Zaremba.
\newblock {OpenAI Gym}, 2016.

\bibitem[Catto(2017)]{Box2D}
Erin Catto.
\newblock A 2d physics engine for games, 2017.
\newblock URL \url{http://box2d.org}.

\bibitem[Coumans and Bai(2016--2017)]{coumans2017pybullet}
Erwin Coumans and Yunfei Bai.
\newblock pybullet, a python module for physics simulation for games, robotics
  and machine learning.
\newblock \url{http://pybullet.org/}, 2016--2017.

\bibitem[Hinton and Van~Camp(1993)]{hinton1993keeping}
Geoffrey~E Hinton and Drew Van~Camp.
\newblock Keeping the neural networks simple by minimizing the description
  length of the weights.
\newblock In \emph{Proceedings of the sixth annual conference on Computational
  learning theory}, pages 5--13. ACM, 1993.

\bibitem[Houthooft et~al.(2016)Houthooft, Chen, Duan, Schulman, De~Turck, and
  Abbeel]{houthooft2016vime}
Rein Houthooft, Xi~Chen, Yan Duan, John Schulman, Filip De~Turck, and Pieter
  Abbeel.
\newblock {VIME}: Variational information maximizing exploration.
\newblock In \emph{Advances in Neural Information Processing Systems}, pages
  1109--1117, 2016.

\bibitem[Lange et~al.(2012)Lange, Riedmiller, and
  Voigtlander]{lange2012autonomous}
Sascha Lange, Martin Riedmiller, and Arne Voigtlander.
\newblock Autonomous reinforcement learning on raw visual input data in a real
  world application.
\newblock In \emph{Neural Networks (IJCNN), The 2012 International Joint
  Conference on}, pages 1--8. IEEE, 2012.

\bibitem[Luck et~al.(2016)Luck, Pajarinen, Berger, Kyrki, and
  Amor]{luck2016sparse}
Kevin~Sebastian Luck, Joni Pajarinen, Erik Berger, Ville Kyrki, and Heni~Ben
  Amor.
\newblock Sparse latent space policy search.
\newblock In \emph{Association for the Advancement of Artificial Intelligence},
  2016.

\bibitem[Mnih et~al.(2016)Mnih, Badia, Mirza, Graves, Lillicrap, Harley,
  Silver, and Kavukcuoglu]{mnih2016asynchronous}
Volodymyr Mnih, Adria~Puigdomenech Badia, Mehdi Mirza, Alex Graves, Timothy
  Lillicrap, Tim Harley, David Silver, and Koray Kavukcuoglu.
\newblock Asynchronous methods for deep reinforcement learning.
\newblock In \emph{International Conference on Machine Learning}, pages
  1928--1937, 2016.

\bibitem[Parr et~al.(2008)Parr, Li, Taylor, Painter-Wakefield, and
  Littman]{parr2008analysis}
Ronald Parr, Lihong Li, Gavin Taylor, Christopher Painter-Wakefield, and
  Michael~L Littman.
\newblock An analysis of linear models, linear value-function approximation,
  and feature selection for reinforcement learning.
\newblock In \emph{Proceedings of the 25th international conference on Machine
  learning}, pages 752--759. ACM, 2008.

\bibitem[Pathak et~al.(2017)Pathak, Agrawal, Efros, and
  Darrell]{pathak2017curiosity}
Deepak Pathak, Pulkit Agrawal, Alexei~A Efros, and Trevor Darrell.
\newblock Curiosity-driven exploration by self-supervised prediction.
\newblock \emph{arXiv preprint arXiv:1705.05363}, 2017.

\bibitem[Schulman et~al.(2017)Schulman, Wolski, Dhariwal, Radford, and
  Klimov]{schulman2017ppo}
John Schulman, Filip Wolski, Prafulla Dhariwal, Alec Radford, and Oleg Klimov.
\newblock Proximal policy optimization algorithms.
\newblock \emph{CoRR}, abs/1707.06347, 2017.
\newblock URL \url{http://arxiv.org/abs/1707.06347}.

\bibitem[Stadie et~al.(2015)Stadie, Levine, and
  Abbeel]{DBLP:journals/corr/StadieLA15}
Bradly~C. Stadie, Sergey Levine, and Pieter Abbeel.
\newblock Incentivizing exploration in reinforcement learning with deep
  predictive models.
\newblock \emph{CoRR}, abs/1507.00814, 2015.
\newblock URL \url{http://arxiv.org/abs/1507.00814}.

\bibitem[Sutton(1996)]{sutton1996generalization}
Richard~S Sutton.
\newblock Generalization in reinforcement learning: Successful examples using
  sparse coarse coding.
\newblock In \emph{Advances in neural information processing systems}, pages
  1038--1044, 1996.

\bibitem[Sutton et~al.(2012)Sutton, Szepesv{\'{a}}ri, Geramifard, and
  Bowling]{sutton2008dyna}
Richard~S. Sutton, Csaba Szepesv{\'{a}}ri, Alborz Geramifard, and Michael
  Bowling.
\newblock Dyna-style planning with linear function approximation and
  prioritized sweeping.
\newblock \emph{CoRR}, abs/1206.3285, 2012.
\newblock URL \url{http://arxiv.org/abs/1206.3285}.

\end{thebibliography}


\section*{Appendix}
\label{sec:appendix}

	\subsection*{Hyperparameters for PPO, IMLE, and VIME}
    
    	We use a standard PPO algorithm adapted from the OpenAI baselines implementation (Table \ref{ppo-hyperp}).
        
        \begin{table}[htb!]
          \caption{PPO Hyperparameters}
          \label{ppo-hyperp}
          \centering
          \begin{tabular}{ll}
            \toprule
            Parameter     	& Value \\
            \midrule
            \# processes 	& 1   	\\
            Epoch steps  	& 2048 	\\
            Entropy coef.   & 0	    \\
            PPO epochs		& 10	\\
            PPO clip		& 0.2 	\\
            PPO batch size	& 64	\\
            Learning rate	& 3e-4	\\
            Discount ($\gamma$) & 0.99 \\
            GAE ($\tau$)	& 0.95	\\
            Horizon			& 1000 (500 in Acrobot) \\
            Steps			& 500000 (250000 in Acrobot and MountainCar) \\
            \bottomrule
          \end{tabular}
        \end{table}
        
        For IMLE and VIME we use hyperparameter settings as close to the original paper as possible (Table \ref{vime-hyperp}).

          \begin{table}[htb!]
          \caption{IMLE and VIME Hyperparameters}
          \label{vime-hyperp}
          \centering
          \begin{tabular}{ll}
            \toprule
            Parameter	     		& Value		\\
            \midrule
            BNN updates per step 	& 500	    \\
            BNN num samples		    & 10	    \\
            BNN batch size		    & 32		\\
            BNN update interval		& 1			\\
            BNN $\eta$				& 0.0001	\\
            Min replay size			& 500		\\
            KL queue length			& 10   		\\
            \bottomrule
          \end{tabular}
          \end{table}

	\subsection*{Proof of that KL divergence is unchanged under linear transformation}

	Given an estimate mean, $\mu_1$, and standard deviation, $\sigma_1$, defined by the output of the dynamics model for the agent at a given time step, $t$. We let the agent take an action $a$ and transition to state $s_{t+1}$. Updating the dynamics model given this new information we have a new output distribution defined by $\mu_2$ and $\sigma_2$. We examine the KL divergence between $p(x) = \mathcal{N}(\mu_1, \sigma_1)$ and $q(x) = \mathcal{N}(\mu_2, \sigma_2)$.
    
    
    Since we assume a network structure such that the value estimate is only a linear transformation of the output of the dynamics model, it too defines a Gaussian distribution. Given parameters $w$ and $b$ of the linear transformation, these transformed distributions, $r(x) = \mathcal{N}(w\mu_1 + b, w^2\sigma_1)$ and $s(x) = \mathcal{N}(w\mu_2 + b, w^2\sigma_2)$ represent the distribution of $Q(s_t, a_t)$ before and after observing $s_{t+1}$. Moreover, the KL divergence of the output of dynamics model is equivalent to the transformed distribution. 
    
    Using that the KL divergence of two multivariate Gaussian distributions is, 
    
    \begin{align}
    	D_{KL}(p, q) &= \int p(x) \frac{\log p(x)}{\log q(x)} dx\\
        			 &= \frac{1}{2} \left( \log\left(\frac{\det \Sigma_2 }{ \det \Sigma_1}\right) + \Tr (\Sigma_2^{-1} \Sigma_1) + (\mu_2 - \mu_1)' \Sigma_2^{-1} (\mu_2 - \mu_1) - N \right)
    \end{align}
    then,

	\begin{align}
    D_{KL}(r, s) &= \frac{1}{2} \bigg( \log\Big(\frac{\det W\Sigma_2W^T }{ \det W\Sigma_1W^T}\Big) + \Tr ((W\Sigma_2W^T)^{-1} (W\Sigma_1W^T)) + \\
    			 & \quad\quad ((W\mu_2 + \bm{b}) - (W\mu_1+\bm{b}))' (W\Sigma_2W^T)^{-1} ((W\mu_2 + \bm{b}) - (W\mu_1 + \bm{b})) - N \bigg) \nonumber \\
                 &= \frac{1}{2} \bigg( \log\Big(\frac{\det \Sigma_2}{\det \Sigma_1}\Big) + \Tr ((W\Sigma_2W^T)^{-1} (W\Sigma_1W^T)) + \\
    			 & \quad\quad ((W\mu_2 + \bm{b}) - (W\mu_1+\bm{b}))' (W\Sigma_2W^T)^{-1} ((W\mu_2 + \bm{b}) - (W\mu_1 + \bm{b})) - N \bigg) \nonumber \\
                 &= \frac{1}{2} \bigg( \log\Big(\frac{\det \Sigma_2}{\det \Sigma_1}\Big) + \Tr ((W^T)^{-1}\Sigma_2^{-1}W^{-1} W\Sigma_1W^T) + \\
    			 & \quad\quad (W\mu_2 - W\mu_1)' (W\Sigma_2W^T)^{-1} (W\mu_2 - W\mu_1) - N \bigg) \nonumber \\ 
                 &= \frac{1}{2} \bigg( \log\Big(\frac{\det \Sigma_2}{\det \Sigma_1}\Big) + \Tr (\Sigma_2^{-1}\Sigma_1) + \\
                 & \quad\quad(\mu_2 - \mu_1)'W^T (W^T)^{-1}\Sigma_2^{-1} W^{-1} W(\mu_2 - \mu_1) - N \bigg) \nonumber \\ 
                 &= \frac{1}{2} \bigg( \log\Big(\frac{\det \Sigma_2}{\det \Sigma_1}\Big) + \Tr (\Sigma_2^{-1}\Sigma_1) + (\mu_2 - \mu_1)'\Sigma_2^{-1}(\mu_2 - \mu_1) - N \bigg) \\ 
                 &= D_{KL}(p, q)
    \end{align}


\end{document}